\documentclass[10pt,twocolumn,letterpaper]{article}

\usepackage{cvpr}              
\definecolor{cvprblue}{rgb}{0.21,0.49,0.74}
\usepackage[pagebackref,breaklinks,colorlinks,allcolors=cvprblue]{hyperref}
\usepackage{multirow}
\usepackage{makecell}
\usepackage[table]{xcolor}
\usepackage{algorithm}
\usepackage{algorithmicx}
\usepackage{algpseudocode}
\def\paperID{*****} 
\def\confName{CVPR}
\def\confYear{2026}

\title{TUDSR: Twice Upsampling-Diffusion for Higher Super-Resolution}

\author{Zhiqiang Wu$^{1}$ \quad Yitong Dong$^{2}$ \quad Xian Wei$^{1}$\thanks{Corresponding author.}
\\
$^{1}$East China Normal University \quad
$^{2}$Zhejiang University \\
\texttt{51265902095@stu.ecnu.edu.cn}
}

\begin{document}
\maketitle
\begin{abstract}
Diffusion-based generative models have achieved remarkable success in real-world image super-resolution (SR). With tiled diffusion techniques, these models can produce high-resolution images that exceed their native-supported resolution. However, the quality of such high-resolution (e.g $2048^2$) outputs often remains extremely poor, primarily due to two factors we consider: the image upsampling ratio (e.g $\times8$) exceeding the model's native-supported upsampling ratio (e.g $\times4$), and the model's native-supported resolution. In practice, training a native high-resolution model requires larger architectures, which incur significant computational overhead and GPU memory costs, making it hard on limited-resource equipment. Thus, we present \textbf{TUDSR}, a \textbf{T}wice \textbf{U}psampling–\textbf{D}iffusion framework for higher SR. The TUDSR framework mainly consists of two stages: the first involves training at $R$-resolution, and the second introduces a looped chunk-based training strategy at $NR$-resolution. Each stage adapts a one-step GAN architecture comprising a generator and a discriminator. Based on SD2.1-base, we develop TUDSR-S, which achieves state-of-the-art performance across multiple benchmarks. Extensive experiments further demonstrate that TUDSR-S generates high-quality images at the resolutions of $1024^2$ and even $2048^2$, significantly outperforming existing approaches. Code is available at \url{https://github.com/wuer5/TUDSR}.
\end{abstract}    
\section{Introduction}
\label{sec:intro}
\begin{figure}[t]
    \centering
    \includegraphics[width=1\linewidth]{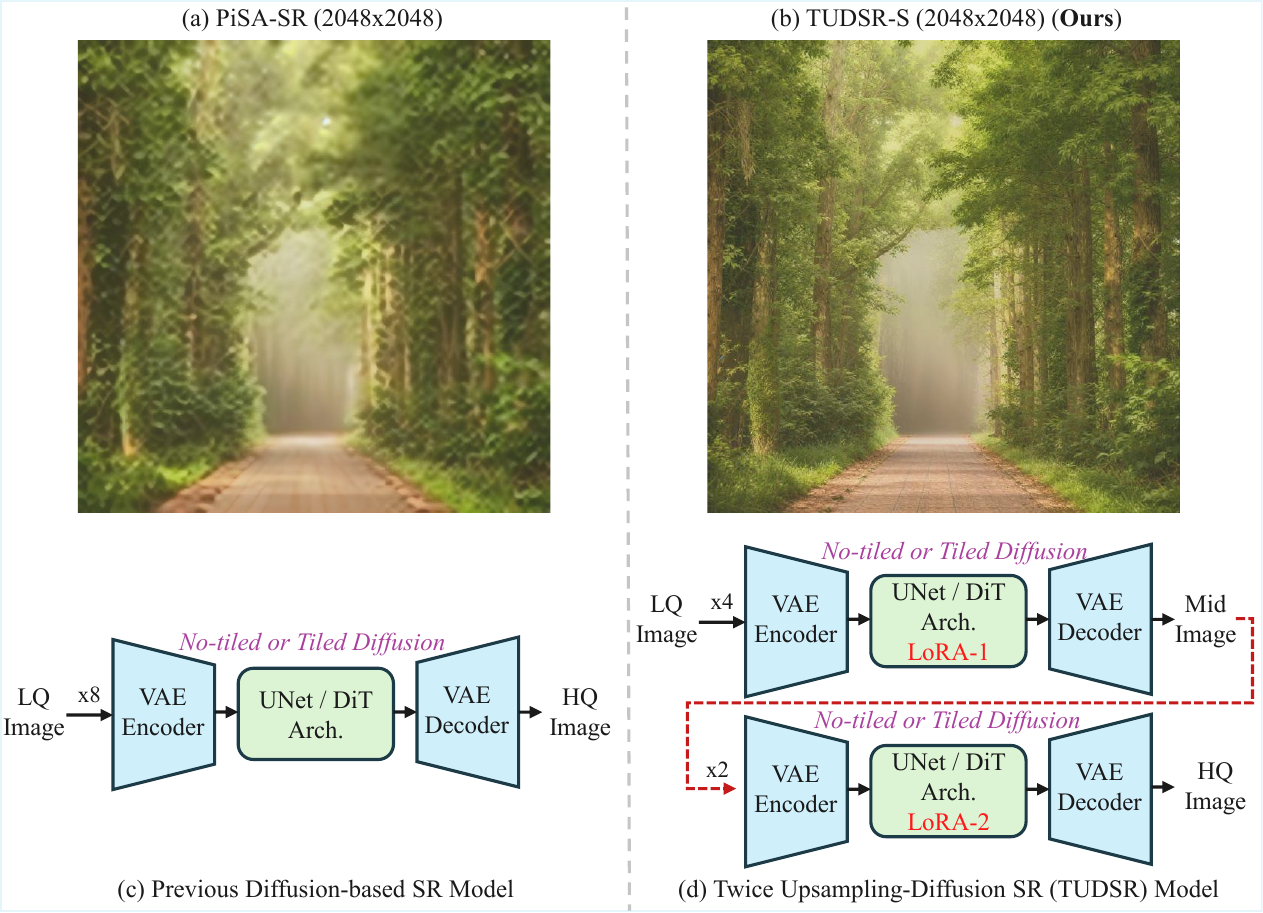}
    \caption{Comparison of (c) Previous Diffusion-based SR Model \vs (d) Twice Upsampling-Diffusion SR (TUDSR) Model. We present a case ($256^2\rightarrow{2048^2}$) from RealLQ250 on (a) PiSA-SR and (b) TUDSR-S (SD2.1-base). Please \textbf{zoom in}.}
    \label{fig:intro}
\end{figure}
Image super-resolution (SR)~\cite{sr1,sr3,sr4,dong2026one} is a fundamental low-level vision task that aims to reconstruct high-quality (HQ) images from their low-quality (LQ) counterparts. Unlike the traditional SR tasks, current SR tasks mainly focus on the real-world scene~\cite{realsr1,realsr2,realesrgan}, where the types of degradations are complex and varying. Such a situation requires the SR model to have strong generalization ability, rather than simply fitting ability for supervised tasks. The recent emergence of large-scale diffusion models~\cite{sd} has revolutionized SR, leveraging powerful generative priors~\cite{diffusionsr, osediff, sinsr, pisasr} to adapt to the real-world degradations.

There are two main types of diffusion-based SR models: multi-step and one-step models. They typically use Stable Diffusion (SD)~\cite{sd} as the foundational model and perform SR downstream tasks via LoRA~\cite{lora} fine-tuning or ControlNet~\cite{controlnet} methods. A common practice among these models is to operate within a native-supported resolution of $512^2$ (\eg \cite{osediff, sinsr, pisasr}). For the images of varying resolution (height and width may be different), the tiled diffusion~\cite{tiled} is introduced during the inference. 

However, the native resolution limit of widely used SD models (\eg SD2.1-base and SD2.1), typically at $512^2$ or $768^2$, severely limits their applicability to high-resolution upsampling tasks such as $128^2\rightarrow1024^2$ or $256^2\rightarrow2048^2$. The required $\times8$ upsampling ratio and the target resolution far exceed the capabilities of these SD models.

One way is to use larger generative models (\eg SD3.5~\cite{sd3} or FLUX.1-dev~\cite{flux}) for training SR tasks with the resolution of $1024^2$. Recent work, FluxSR~\cite{fluxsr}, has employed FLUX.1-dev as the generative model for SR tasks. However, this way may result in significant computational overhead and GPU memory costs, making the higher SR difficult to train and deploy on limited devices.

For a higher SR task, it is common practice to first upsample a low-resolution image (\eg $256^2$) to a high-resolution image (\eg $2048^2$) before applying an SR model. However, direct upsampling at high ratios often exceeds the capacity of existing SR models. 
To address this limitation, we propose \textbf{TUDSR}, a \textbf{T}wice \textbf{U}psampling–\textbf{D}iffusion framework. Specifically, we employ two LoRA adapters trained separately within the same generative model. We divide the training process into two stages. The first stage involves training a LoRA adapter at $R$-resolution. In the second stage, we keep the first-stage LoRA SR model fixed and use its outputs as inputs to train the second-stage LoRA adapter.
Before the second-stage training, the input is $\times N$ upsampled, then we adapt a for-loop chunked training strategy to reduce GPU memory costs. During inference, both LoRA adapters are utilized sequentially: the input is first $\times M$ upsampled and processed by the first-stage LoRA SR model; the output is then $\times N$ upsampled again and passed through the second-stage LoRA SR model to produce the final image. During inference, we only need to switch between two LoRA adapters (lower GPU memory usage) in the backbone network of a generative model, \textbf{\textit{without}} initializing or loading the backbone network for both LoRAs. Note that $\times MN$ is your target upsampling ratio. 

As shown in \cref{fig:intro}, we provide a comparison between our TUDSR framework and traditional SR models, and also present a comparison (PiSA-SR~\cite{pisasr} \vs TUDSR-S) at the resolution of $2048^2$. PiSA-SR produces low-quality images, which appear blurry and lack detail. On the contrary, our TUDSR-S generates forest scenes with clean, rich details, fully leveraging the SD model's image priors.

Under the TUDSR framework, we circumvent the challenges of one-step high-ratio upsampling by decomposing the process into a two-stage upsampling–diffusion pipeline, thereby fully leveraging the generative model's high-resolution generation capability. 

\textbf{The main contributions are as follows:}
\begin{itemize}
    \item To the best of our knowledge, our TUDSR is the first to generate high-quality, high-resolution (\eg $2048^2$) images from a native low-resolution (\eg $512^2$) generative model, making it deployable on resource-constrained devices without a larger model architecture.
    \item Our TUDSR framework provides a training and inference strategy for higher SR tasks. For example, TUDSR can be extended to larger-scale generative models (\eg FLUX.1-dev) for higher SR ($4096^2$).
    \item Based on SD2.1-base, we instantiate TUDSR-S, achieving state-of-the-art results across multiple datasets and metrics for both $\times4$ and $\times8$ SR tasks. In particular, it shows the best performance on the $\times8$ SR task.
\end{itemize}

\section{Related Work}
\label{sec:related}
\subsection{Deep Learning-Based Super-Resolution}
Early SR research trained networks on bicubically downsampled image pairs, but these models failed to generalize due to complex real-world degradations. Collecting real LQ-HQ pairs is a solution, but the cost is prohibitive. Thus, simulating realistic degradation is still a key research direction. Real-ESRGAN~\cite{realesrgan} introduces a two-stage degradation process and adversarial training~\cite{gan}, improving perceptual quality significantly. However, these full-parameter training methods remain limited by their training data and cannot fully address diverse real-world degradations.

\subsection{Diffusion-Based Super-Resolution}
Early diffusion-based SR methods often oversimplified real-world degradations and overlooked the value of powerful image priors. With the rise of large-scale diffusion models like SD~\cite{sd}, recent works leverage their strong pre-trained priors for SR. For example, StableSR~\cite{stablesr} conditions on the LR input; DiffBIR~\cite{diffbir} uses a two-stage degradation removal and enhancement pipeline; and SeeSR~\cite{seesr} employs a degradation-aware prompt extractor. These approaches typically rely on standard DDPM~\cite{ddpm} training and multi-step sampling, resulting in two key drawbacks: slow inference and potential low fidelity or unrealistic outputs, limiting their suitability for fidelity-critical SR.

To address these issues, recent one-step models have emerged. Methods such as SinSR~\cite{sinsr} and OSEDiff~\cite{osediff} use distillation to compress multi-step models into one-step models. SinSR introduces a consistency-preserving loss to shorten the diffusion trajectory, while OSEDiff employs Variational Score Distillation (VSD) to distill the generalization capability of Stable Diffusion. PiSA-SR~\cite{pisasr} learns two LoRA modules for an SD model to achieve improved and adjustable SR at both pixel and semantic levels. InvSR~\cite{invsr} designs a Partial Noise Prediction strategy to construct an intermediate state of the SD model as the sampling start point, enabling either multi-step or one-step prediction by configuring the number of intermediate timesteps.

\subsection{Higher Super-Resolution}
Although these diffusion-based models can generate high-resolution images via tiled diffusion~\cite{tiled}, they are inherently constrained by their SD model, which typically supports resolutions like $512^2$ or $768^2$. For high-resolution (\eg $2048^2$) outputs requiring large upscaling factors (\eg $\times 8$), these models (\eg \cite{osediff,pisasr,sinsr,seesr,diffbir,stablesr,invsr,resshift}) struggle to produce high-quality results. While scaling to larger generative models (\eg FLUX.1-dev~\cite{flux}) is possible, the associated training computational cost would be immense. Thus, this paper will focus specifically on the upscaling factor to construct a higher SR framework.

\begin{figure*}[t]
    \centering
    \includegraphics[width=0.96\linewidth]{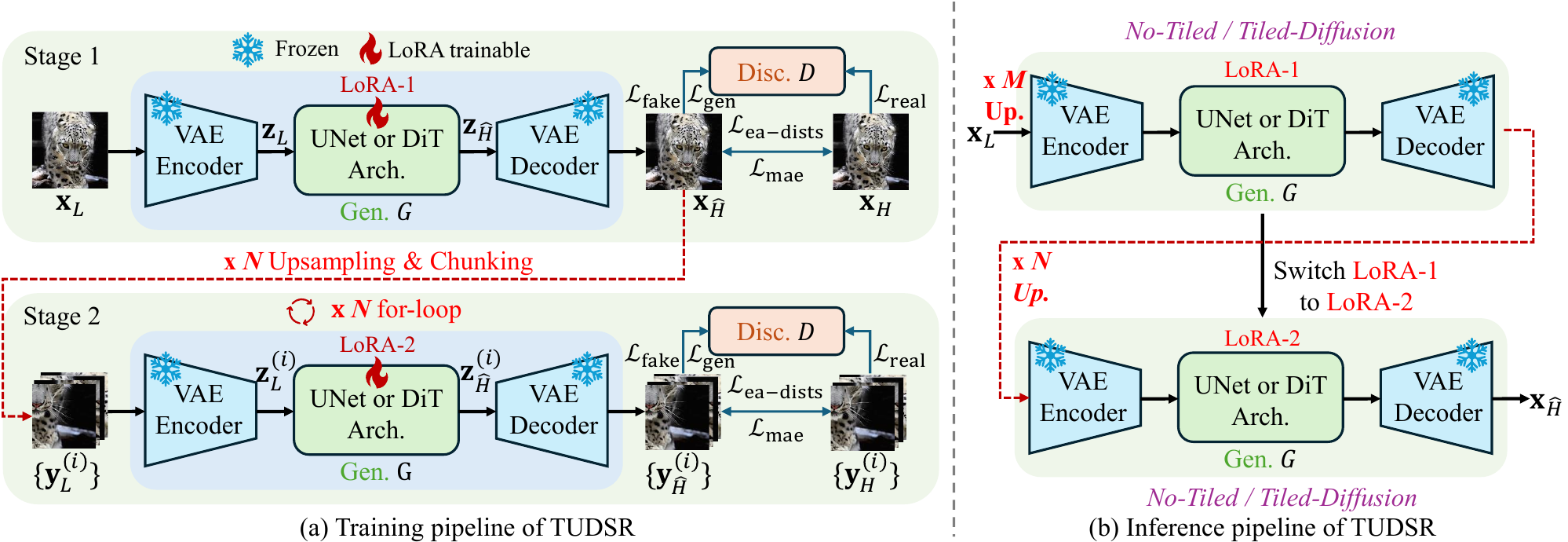}
    \caption{Illustration of the (a) training and (b) inference pipelines of TUDSR. In stage 1, we train a one-step LoRA at $R$-resolution. In stage 2, we freeze the first-stage LoRA SR model to generate an intermediate image, which is then $\times N$ upsampled. Subsequently, we train a second-stage LoRA to process the upsampled image using a for-loop chunk-wise training strategy, where gradients are backpropagated per chunk. Notably, both stages employ a one-step training approach based on a GAN architecture~\cite {gan}. The inference process employs the two LoRA adapters in sequence: the input first undergoes $\times M$ upsampling by the first-stage LoRA SR model, followed by $\times N$ upsampling by the second-stage model, yielding the final output.}
    \vspace{-2mm}
    \label{fig:arch}
\end{figure*}
\section{Methodology}
\subsection{Diffusion-based SR Modeling}
Given an LQ image $\mathbf{x}_{L}$ with complex and varying real-world degradations, we aim to generate the corresponding HQ image $\mathbf{x}_{H}$. Our task mainly focuses on the SD model, which consists of a VAE encoder~\cite{vae} $\mathcal{E}$ for compressing images into latent space, a backbone for denosing $\mathcal{B}$, and a VAE decoder~\cite{vae} $\mathcal{D}$ for constructing the latent representation into the image. The problem is formulated as 
\begin{equation}
    \mathbf{x}_{\widehat{H}} = \mathcal{D}(\mathbf{z}_{\widehat{H}}),~\mathbf{z}_{\widehat{H}} = \operatorname{f}_{G}(\mathbf{z}_L),~
\mathbf{z}_L = \mathcal{E}(\mathbf{x}_{L}), 
\end{equation}
where $\operatorname{f}_G$ denotes the denoising process.

\subsection{Architeture of TUDSR}
We propose TUDSR, a Twice Upsampling-Diffusion SR framework, as illustrated in \cref{fig:arch}. TUDSR uses two-stage training, where each stage employs a GAN~\cite{gan} architecture (Generator $G$ and Discriminator $D$).
\begin{itemize}
    \item \textbf{Generator $G$:} We employ a pre-trained generative model (training via LoRA) as the generator.
    \item \textbf{Discriminator $D$:} We employ a pre-trained DINOv3-ViT-B~\cite{dinov3} model as the feature extractor within a multi-level discriminator, which incorporates multi-level discriminator heads for adversarial learning~\cite{gan}.
\end{itemize}
\subsection{Rethinkings of GAN Architecture}
In traditional GAN training, the difficulties primarily stem from two factors: (1) the common training paradigm of mapping noise to HQ images involves high task complexity and diversity; (2) training both the generator and discriminator from scratch makes it challenging to maintain a balanced optimization dynamic. However, in GANs for SR, the task is simplified to mapping from LQ to HQ images. Furthermore, since designing and training both a discriminator and a generator from scratch is complex and prone to imbalance, we adapt a pre-trained diffusion model as the generator and fine-tune it using LoRA. On the other hand, we leverage a pre-trained DINOv3-ViT-B~\cite{dinov3} to extract multi-level features and train multi-level discriminator heads from scratch. As a result, both the generator and the discriminator are equipped with strong prior knowledge, while the number of trainable parameters remains relatively small. This leads to significantly enhanced stability in GAN-based SR training. Moreover, the GAN framework enables the generation of higher-quality images with superior detail rendition compared to other architectures.
\subsection{GAN Generator $G$}
In our TUDSR framework, we see a pre-trained generative model (\eg SD2.1-base) with LoRA fine-tuning as the generator. Given a LQ image $\mathbf{x}_{L}$, we define the generator as the function $G$
\begin{equation}
    \label{eq2}
     G(\mathbf{x}_{L},t,c) = \mathbf{x}_{\widehat{H}},~\text{where}
    \begin{cases}
    \mathbf{z}_{L}=\mathcal{E}(\mathbf{x}_{L})
     \\ \mathbf{z}_{\widehat{H}} = \frac{\mathbf{z}_{L} - \sqrt{1-\bar{\alpha}_t}  \cdot \mathcal{B}(\mathbf{z}_{L}, t, c)}{\sqrt{\bar{\alpha}_t}}
     \\ \mathbf{z}_{\widehat{H}}=\mathcal{D}(\mathbf{x}_{\widehat{H}})
    \end{cases}
\end{equation}
Note that $t$ is the fixed timestep for onestep denoising and $c$ is the prompt condition.

\begin{figure}
    \centering
    \includegraphics[width=0.96\linewidth]{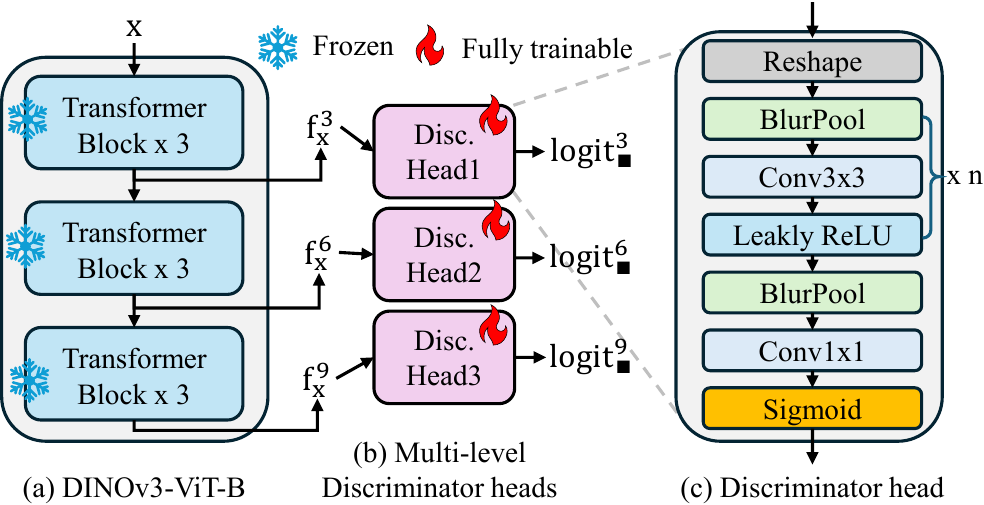}
    \caption{Illustration of Discriminator $D$ (DINOv3-ViT-B + Multi-level Discriminator Heads). Note that $\operatorname{BlurPool}$~\cite{bluepool} is a low-pass filter used for anti-aliasing, which is a commonly used method in the design of GAN discriminators.}
    \label{fig:disc}
\end{figure}
\subsection{GAN Discriminator $D$}
The GAN discriminator consists of a frozen pre-trained backbone DINOv3-ViT-B for feature extraction and fully trainable multi-level discriminator heads to predict the discrimination logits.
\subsubsection{DINOv3-ViT-B Features}
The DINOv3-ViT-B ($86$M parameters) adapts the standard ViT~\cite{vit} architecture for pre-training on large-scale datasets, which contain powerful image priors. It is mainly composed of $12$ layers of transform blocks, and a large body of literature~\cite{amir2021deep,xia2024vit,chen2025frequency,hernandez2025vision} shows that the shallow and middle layers contain detailed information, while the higher layers primarily encode global semantic information. For SR tasks, LQ images themselves contain global semantic information, but lack detailed information. Therefore, we use the features in $3$, $6$, and $9$ layers for detailed discrimination here.
Given the input image $\mathbf{x}$, we define the extraction process of DINOv3-ViT-B as
\begin{equation}
    \operatorname{DINOv3ViTB}(\mathbf{\mathbf{x}}, \{3,6,9\}) = \{\operatorname{f}_\mathbf{\mathbf{x}}^3, \operatorname{f}_\mathbf{\mathbf{x}}^6, \operatorname{f}_\mathbf{\mathbf{x}}^9\}.
\end{equation}
\subsubsection{Multi-level Discriminator Heads}
Once we obtain the features from DINOv3-ViT-B, we design multi-level discriminator heads $\operatorname{DHead}^l$ for discrimination prediction, as shown in \cref{fig:disc}. We define the discriminator as the function $D$ where
\begin{equation}
     D(\mathbf{x})= \{\operatorname{DHead}^l(\operatorname{f}_{\operatorname{\mathbf{x}}}^l)\}=\{\operatorname{logits}^l_{\Box}\},~l=3,6,9.
\end{equation}
\subsection{Training Objective}
\subsubsection{Structural Perception Loss}
Although LPIPS~\cite{lpips} has been widely adopted for structural perception, several studies and our own experiments show that it can introduce artifacts, particularly during diffusion-based GAN training. To address this, we employ dists as our structural perception loss. Unlike LPIPS, which computes a weighted L2 distance between multi-level feature maps from a pre-trained network, dists measures the discrepancy in first-order (mean) and second-order (covariance) statistics captured by these feature maps. This approach exhibits greater robustness to geometric distortions and luminance variations, reduces the likelihood of artifacts, and aligns more closely with human visual perception. Furthermore, as demonstrated in \cite{eadists}, edge details are also crucial for structural perception. Following \cite{eadists}, we integrate an edge-aware version of dists to enhance performance:
\begin{equation}
    \mathcal{L}_{\operatorname{ea-dists}} = \operatorname{dists}(\mathbf{x}_{H}, \mathbf{x}_{\widehat{H}}) + \operatorname{dists}(\mathcal{S}(\mathbf{x}_{H}), \mathcal{S}(\mathbf{x}_{\widehat{H}})),
\end{equation}
where $\mathcal{S}(\cdot)$ denotes the Sobel operator to extract the edge information of images. 

\subsubsection{GAN Generator Loss}
The GAN generator loss is used to update the generator's parameters. We freeze the discriminator $D$ to obtain the generative logits: $D(\mathbf{x}_{\widehat{H}})=\{\operatorname{logit}^l_{\operatorname{gen}}\}$ and employ the Binary Cross-Entropy (BCE):

\begin{equation}
    \mathcal{L}_{\operatorname{gen}} =  \frac{1}{3}\sum^{\{3,6,9\}}_{l} \operatorname{BCE}(\operatorname{label}_{\operatorname{gen}}, \operatorname{logit}^l_{\operatorname{gen}}),
\end{equation}
where $\operatorname{label}_{\operatorname{gen}}$ is the soft label and set to $0.8$.
\subsubsection{GAN Discriminator Loss}
The GAN generator loss is used to discriminate between real and fake images to update the discriminator's parameters. We detach $\mathbf{x}_{\widehat{H}}$ and obtain the fake logits: $D(\operatorname{Detach}(\mathbf{x}_{\widehat{H}}))=\{\operatorname{logit}^l_{\operatorname{fake}}\}$ and real logits: $D(\mathbf{x}_{H})=\{\operatorname{logit}^l_{\operatorname{real}}\}$.
We also use the Binary Cross-Entropy (BCE):
\begin{equation}
    \begin{split}
        &\mathcal{L}_{\operatorname{real}} =  \frac{1}{3}\sum^{\{3,6,9\}}_{l} \operatorname{BCE}(\operatorname{label}_{\operatorname{real}}, \operatorname{logit}^l_{\operatorname{real}}),
        \\
        &\mathcal{L}_{\operatorname{fake}} =  \frac{1}{3}\sum^{\{3,6,9\}}_{l} \operatorname{BCE}(\operatorname{label}_{\operatorname{fake}}, \operatorname{logit}^l_{\operatorname{fake}}),
    \end{split}
\end{equation}
where $\operatorname{label}_{\operatorname{fake}}$ and $\operatorname{label}_{\operatorname{real}}$ are set to $0$ and $0.8$.
\subsubsection{Total Training Objective}
\label{obj}
For the generator, we optimize the LoRA parameters of the generative model via the loss:
\begin{equation}
    \mathcal{L}_{\operatorname{G}} = \lambda_1\mathcal{L}_{\operatorname{ea-dists}} + \lambda_2\mathcal{L}_{\operatorname{gen}}+\lambda_3\mathcal{L}_{\operatorname{mae}},
\end{equation}
where $\lambda_1=5$, $\lambda_2=0.5$, and $\lambda_3=0.5$ in this paper.

For the discriminator, we optimize all the parameters of the multi-level discriminator heads via the loss:
\begin{equation}
    \mathcal{L}_{\operatorname{D}} = \lambda_2(\mathcal{L}_{\operatorname{real}} +\mathcal{L}_{\operatorname{fake}}).
\end{equation}
\subsection{Stage 1: $R$-resolution Training}
In the first stage, we train a LoRA of the generative model as $R$-resolution (\eg $512$).
We optimize the LoRA parameters $\theta_1$ of a generative model and the parameters $\phi_1$ of the multi-level discriminator heads. Given the LQ image $\mathbf{x}_L$ with the resolution of $R^2$, the stage-1 training is described in \cref{alg:stage1}.
\begin{algorithm}
\caption{Stage 1 Training}
\label{alg:stage1}
\begin{algorithmic}[0] 
    \Require $\mathbf{x}_L$, $\mathbf{x}_H$, $t_1$, $c$, $G_{\theta_1}$, $D_{\phi_1}$
    \State $\mathbf{x}_{\widehat{H}} \gets G_{\theta_1}(\mathbf{x}_L,t_1,c)$
    \State $\{\operatorname{logit}^l_{\operatorname{gen}}\} \gets D_{\phi_1}^\ast(\mathbf{x}_{\widehat{H}})$ \Comment{$^\ast$: freeze $D_{\phi_1}$, for $G$}
    \State $\mathcal{L}_{\operatorname{G}}^{\theta_1} \gets \lambda_1\mathcal{L}_{\operatorname{ea-dists}}(\mathbf{x}_{\widehat{H}},\mathbf{x}_{H}) + \lambda_2\mathcal{L}_{\operatorname{gen}}(\{\operatorname{logit}^l_{\operatorname{gen}}\})$\\~~~~~~~~~~~~$+\lambda_3\mathcal{L}_{\operatorname{mae}}(\mathbf{x}_{\widehat{H}},\mathbf{x}_{H})$
    \State $\theta_1 \gets\operatorname{Backward}(\mathcal{L}_{\operatorname{G}}^{\theta_1})$ \Comment{Update $\theta_1$}
    \State $\{\operatorname{logit}^l_{\operatorname{fake}}\} \gets D_{\phi_1}(\operatorname{Detach}
    (\mathbf{x}_{\widehat{H}}))$ \Comment{For real image}
    \State $\{\operatorname{logit}^l_{\operatorname{real}}\} \gets D_{\phi_1}(\mathbf{x}_{H})$ 
    \Comment{For fake image}
    \State $\mathcal{L}_{\operatorname{D}}^{\phi_1} \gets \lambda_2(\mathcal{L}_{\operatorname{fake}}(\{\operatorname{logit}^l_{\operatorname{fake}}\})+\mathcal{L}_{\operatorname{real}}(\{\operatorname{logit}^l_{\operatorname{real}}\}))$
    \State $\phi_1 \gets\operatorname{Backward}(\mathcal{L}_{\operatorname{D}}^{\phi_1})$ \Comment{Update $\phi_1$}
\end{algorithmic}
\end{algorithm}

\subsection{Stage 2: $NR$-resolution Training}
In the second stage, we freeze the LoRA of the first stage. Given the LQ image $\mathbf{x}_L$ with the resolution of $R^2$, we freeze $G_{\theta_1}$ and obtain the intermediate image
$\mathbf{m} = G_{\theta_1}(\mathbf{x}_L, t_1, c)$.
Then, we use the bilinear interpolation to $\times N$ upsample $\mathbf{m}$:
\begin{equation}
    \label{eq11}
    \mathbf{y}_L = \operatorname{Upsampling}(\mathbf{m}, N).
\end{equation}
However, directly training the second LoRA at $NR$-resolution introduces two main issues: (1) $NR$-resolution exceeds the model's native-supported resolution (\ie $R$); (2) gradient computation at $NR$-resolution consumes substantial GPU memory.
We adapt a for-loop chunked training strategy and perform gradient updates per chunk:
\begin{equation}
    \{\mathbf{y}_L^{(i)}\} = \operatorname{Chunking}(\mathbf{y}_L,R),~i\in\{1,\cdots,N^2\}.
\end{equation}
Note that $\operatorname{Chunking}$ divides $\mathbf{y}_L$ into chunks from left to right and from top to bottom without overlap.

Given the HQ chunked image $\{\mathbf{y}_H^{(i)}\}$, we optimize the LoRA parameter $\theta_2$ of the generative model and the multi-level discriminator head parameters $\phi_2$. 
The stage-2 training is described in \Cref{alg:stage2}.

\begin{algorithm}
\caption{Stage 2 Training}
\label{alg:stage2}
\begin{algorithmic}[0] 
    \Require $\mathbf{x}_L$, ${\mathbf{y}_H^{(i)}}$, $t_1$, $t_2$, $c$, $G_{\theta_1}$, $G_{\theta_2}$, $D_{\phi_2}$, $R$, $l$, $N$
    \State $\mathbf{m} \gets G_{\theta_1}^\ast(\mathbf{x}_L,t_1,c)$ 
    \Comment{$^\ast$: freeze $G_{\theta_1}$}
    \State $\mathbf{y}_L \gets \operatorname{Upsampling}(\mathbf{m}, N)$ \Comment{$\times N$ upsampling}
    \State $\{\mathbf{y}_L^{(i)}\} \gets \operatorname{Chunking}(\mathbf{y}_L,R),~i\in\{1,\cdots,N^2\}$
    \For{$i = 1$ to $N^2$}
    \Comment{For each chunk}
        \State $\mathbf{y}_{\widehat{H}}^{(i)} = G_{\theta_2}(\mathbf{y}_L^{(i)},t_2,c)$
        \State $\{\operatorname{logit}^l_{\operatorname{gen}}\} \gets D_{\phi_2}^\ast(\mathbf{y}_{\widehat{H}}^{(i)})$ \Comment{$^\ast$: freeze $D_{\phi_2}$, for $G$}
        \State $\mathcal{L}_{\operatorname{G}}^{\theta_1} \gets \lambda_1\mathcal{L}_{\operatorname{ea-dists}}(\mathbf{y}_{\widehat{H}}^{(i)},\mathbf{y}_{H}^{(i)}) + \lambda_2\mathcal{L}_{\operatorname{gen}}(\{\operatorname{logit}^l_{\operatorname{gen}}\})$\\~~~~~~~~~~~~~~~~~~$+\lambda_3\mathcal{L}_{\operatorname{mae}}(\mathbf{y}_{\widehat{H}}^{(i)},\mathbf{y}_{H}^{(i)})$ 
        \State $\theta_2 \gets\operatorname{Backward}(\mathcal{L}_{\operatorname{G}}^{\theta_2})$ \Comment{Update $\theta_2$}
        \State $\{\operatorname{logit}^l_{\operatorname{fake}}\} \gets D_{\phi_2}(\operatorname{Detach}
        (\mathbf{y}_{\widehat{H}}^{(i)}))$ \Comment{For real image}
        \State $\{\operatorname{logit}^l_{\operatorname{real}}\}\gets D_{\phi_2}(\mathbf{y}_{H}^{(i)})$ 
        \Comment{For fake image}
        \State $\mathcal{L}_{\operatorname{D}}^{\phi_2} \gets \lambda_2(\mathcal{L}_{\operatorname{fake}}(\{\operatorname{logit}^l_{\operatorname{fake}}\})+\mathcal{L}_{\operatorname{real}}(\{\operatorname{logit}^l_{\operatorname{real}}\}))$
        \State $\phi_2 \gets\operatorname{Backward}(\mathcal{L}_{\operatorname{D}}^{\phi_2})$ \Comment{Update $\phi_2$}
    \EndFor
\end{algorithmic}
\end{algorithm}
\section{Experiment}
\subsection{Experimental Settings}

\textbf{Training settings.}
We adopt the experimental setup from SeeSR~\cite{seesr}, using the LSDIR~\cite{lsdir} dataset along with the first 10,000 facial images from FFHQ~\cite{ffhq}. LQ-HQ pairs are synthesized via the standard Real-ESRGAN degradation pipeline~\cite{realesrgan}. Within TUDSR, we initialize TUDSR-S from SD2.1-base, setting the rank of UNet LoRA to 32 for both stages. Training uses AdamW~\cite{adamw} with a learning rate of $5 \times 10^{-5}$, batch size of 1, and 4 gradient accumulation steps. We set $N=2$ in \cref{eq11}, $t_1=200$ in \cref{alg:stage1}, and $t_2=50$ in \cref{alg:stage2}. Stage 1 and stage 2 are trained for 5,100 and 3,500 steps, respectively, on 4 RTX 4090 GPUs.

\noindent\textbf{Test Datasets.}
We evaluate on four real-world datasets: RealSR~\cite{realsr}, DrealSR~\cite{drealsr}, RealLQ250~\cite{reallq250}, and RealLR200~\cite{seesr}. RealSR and DrealSR contain 100 and 93 images at $128^2$ with corresponding GT at $512^2$, respectively; RealLQ250 consists of 250 $256^2$ images; RealLR200 includes 200 images of varying resolutions.

\noindent\textbf{Test Metrics.}
As PSNR and SSIM~\cite{ssim} no longer accurately reflect perceptual quality, we adopt model-based metrics: reference metrics LPIPS~\cite{lpips} and FID~\cite{fid}; non-reference metrics CLIPIQA~\cite{clipiqa}, CLIPIQA+~\cite{clipiqa}, NIMA~\cite{nima}, NIQE~\cite{niqe}, LIQE~\cite{liqe}, MUSIQ~\cite{musiq}, and MANIQA~\cite{maniqa}.

\subsection{Comparison Results with Others}
We compare TUDSR-S against state-of-the-art diffusion-based multi-step (StableSR~\cite{stablesr}, DiffBIR~\cite{diffbir}, SeeSR~\cite{seesr}, ResShift~\cite{resshift}) and one-step SR models (SinSR~\cite{sinsr}, OSEDiff~\cite{osediff}, PiSA-SR~\cite{pisasr}, InvSR~\cite{invsr}) on $\times4$ and $\times8$ tasks. Tiled diffusion is used for inputs exceeding 512 pixels. $\times8$ experiments on RealLR200 are omitted due to OOM errors. Given the high inference cost of multi-step models (over 10 minutes per $\times8$ image), we restrict $\times8$ comparisons to one-step models only.

\begin{table*}[h]
  \caption{Quantitative comparisons ($\times 4$) with state-of-the-art multi-step and one-step models on four real-world benchmark datasets. Note that TUDSR-S denotes \texttt{M2N2} twice upsampling-diffusion in the table.}
  \vspace{-3mm}
  \label{tab:cmp4}
  \resizebox{\textwidth}{!}{ 
  \centering
    \begin{tabular}{c|l|ccccc|cccccc}
    \toprule
    Datasets & Metrics & RealESRGAN~\cite{realesrgan} & \makecell{StableSR~\cite{stablesr}} & \makecell{SeeSR~\cite{seesr}} & \makecell{DiffBIR~\cite{diffbir}} & \makecell{ResShift~\cite{resshift}} & \makecell{OSEDiff~\cite{osediff}} & \makecell{PiSASR~\cite{pisasr}} & \makecell{InvSR~\cite{invsr}} & \makecell{SinSR~\cite{sinsr}} & \cellcolor{gray!20}\makecell{TUDSR-S} \\
    \midrule
    \multirow{9}{*}{RealSR} 
     & LPIPS$\downarrow$ & 0.2710 & \textbf{0.2604} & 0.3007 & 0.3470 & 0.3159 & 0.2921 & \underline{0.2672} & 0.2871 & 0.3210 & \cellcolor{gray!20}0.3217 \\
     & FID$\downarrow$ & 135.15 & 132.09 & 125.51 & 134.59 & 149.65 & \underline{123.50} & 124.19 & 138.85 & 136.78 & \cellcolor{gray!20}\textbf{111.42} \\
     & CLIPIQA$\uparrow$ & 0.4490 & 0.5426 & 0.6699 & \textbf{0.6960} & 0.5505 & 0.6693 & 0.6699 & 0.6785 & 0.6156 & \cellcolor{gray!20}\underline{0.6846} \\
     & CLIPIQA+$\uparrow$ & 0.5841 & 0.6150 & 0.6909 & \underline{0.6989} & 0.5451 & 0.6964 & 0.6957 & 0.6880 & 0.5370 & \cellcolor{gray!20}\textbf{0.7135} \\
     & NIMA$\uparrow$ & 4.6551 & 4.6767 & \underline{4.9191} & 4.9159 & 4.7554 & 4.8951 & 4.8953 & \textbf{5.0946} & 4.6643 & \cellcolor{gray!20}4.8676 \\
     & NIQE$\downarrow$ & 5.7960 & 6.6231 & \underline{5.3984} & 5.4992 & 6.8833 & 5.6474 & 5.5057 & 5.6222 & 6.2998 & \cellcolor{gray!20}\textbf{4.7149} \\
     & LIQE$\uparrow$ & 3.3571 & 3.2578 & \underline{4.1354} & 4.0261 & 3.1859 & 4.0690 & 4.0989 & 4.0392 & 3.1466 & \cellcolor{gray!20}\textbf{4.3738} \\
     & MUSIQ$\uparrow$ & 60.3657 & 61.8058 & 69.8165 & 68.3462 & 60.2181 & 69.0896 & \underline{70.1492} & 68.5372 & 60.4204 & \cellcolor{gray!20}\textbf{70.2406} \\
     & MANIQA$\uparrow$ & 0.5492 & 0.5952 & 0.6445 & 0.6540 & 0.5388 & 0.6331 & 0.6552 & \underline{0.6628} & 0.5391 & \cellcolor{gray!20}\textbf{0.6786} \\
    \midrule

    \multirow{9}{*}{DrealSR} 
     & LPIPS$\downarrow$ & \underline{0.2819} & \textbf{0.2698} & 0.3174 & 0.4520 & 0.3526 & 0.2968 & 0.2960 & 0.3538 & 0.3674 & \cellcolor{gray!20}0.3372 \\
     & FID$\downarrow$ & 147.80 & 151.27 & 147.53 & 177.06 & 176.70 & 135.29 & \textbf{130.43} & 171.40 & 171.88 & \cellcolor{gray!20}\underline{133.74} \\
     & CLIPIQA$\uparrow$ & 0.4517 & 0.4907 & 0.6913 & 0.6859 & 0.5410 & 0.6963 & \underline{0.6974} & \textbf{0.7132} & 0.6348 & \cellcolor{gray!20}0.6944 \\
     & CLIPIQA+$\uparrow$ & 0.5544 & 0.5347 & 0.6794 & 0.6828 & 0.5157 & 0.6825 & 0.6920 & \underline{0.6832} & 0.5400 & \cellcolor{gray!20}\textbf{0.6921} \\
     & NIMA$\uparrow$ & 4.3261 & 4.2136 & 4.6945 & \underline{4.7847} & 4.4405 & 4.6766 & 4.6250 & \textbf{4.8566} & 4.4639 & \cellcolor{gray!20}4.7272 \\
     & NIQE$\downarrow$ & 6.6927 & 7.5488 & 6.4136 & 6.2409 & 7.8693 & 6.4904 & 6.1759 & \underline{5.9917} & 7.1422 & \cellcolor{gray!20}\textbf{5.5427} \\
     & LIQE$\uparrow$ & 2.9259 & 2.4349 & \underline{4.1268} & 3.8930 & 2.7968 & 3.9371 & 4.0440 & 4.0557 & 3.0514 & \cellcolor{gray!20}\textbf{4.1769} \\
     & MUSIQ$\uparrow$ & 54.2721 & 51.3635 & 65.0935 & 65.6585 & 52.3726 & 64.6537 & \underline{66.1094} & 65.9956 & 54.9825 & \cellcolor{gray!20}\textbf{66.3650} \\
     & MANIQA$\uparrow$ & 0.4899 & 0.4969 & 0.6043 & 0.6279 & 0.4750 & 0.5895 & 0.6146 & \underline{0.6302} & 0.4855 & \cellcolor{gray!20}\textbf{0.6312} \\
    \midrule

    \multirow{7}{*}{RealLQ250} 
     & CLIPIQA$\uparrow$ & 0.5434 & 0.5150 & 0.7132 & \underline{0.7255} & 0.4734 & 0.6995 & 0.7095 & 0.6628 & 0.6998 & \cellcolor{gray!20}\textbf{0.7315} \\
     & CLIPIQA+$\uparrow$ & 0.6117 & 0.5811 & 0.7142 & \underline{0.7213} & 0.4642 & 0.7017 & 0.7160 & 0.6722 & 0.5919 & \cellcolor{gray!20}\textbf{0.7297} \\
     & NIMA$\uparrow$ & 5.2554 & 5.0700 & 5.3863 & \textbf{5.4922} & 5.0243 & 5.2364 & 5.2429 & \underline{5.4401} & 5.1938 & \cellcolor{gray!20}5.3268 \\
     & NIQE$\downarrow$ & 4.1292 & 4.6345 & 3.9832 & \textbf{3.5608} & 4.8476 & 3.9127 & \underline{3.8751} & 4.4098 & 5.7974 & \cellcolor{gray!20}3.9440 \\
     & LIQE$\uparrow$ & 3.3410 & 2.7533 & \underline{4.1336} & 4.0874 & 2.4609 & 3.8610 & 3.9813 & 3.7113 & 3.2465 & \cellcolor{gray!20}\textbf{4.3017} \\
     & MUSIQ$\uparrow$ & 62.5169 & 57.1341 & \underline{71.1218} & 70.3687 & 57.7724 & 69.6786 & 71.0710 & 65.8212 & 63.8548 & \cellcolor{gray!20}\textbf{71.9250} \\
     & MANIQA$\uparrow$ & 0.5288 & 0.5203 & 0.6204 & \underline{0.6232} & 0.4657 & 0.5928 & 0.6157 & 0.5914 & 0.5178 & \cellcolor{gray!20}\textbf{0.6329} \\
    \midrule
    \multirow{7}{*}{RealLQ200} 
     & CLIPIQA$\uparrow$ & 0.5409 & 0.5731 & 0.6959 & \underline{0.7222} & 0.4942 & 0.7008 & 0.7125 & 0.6830 & 0.6615 & \cellcolor{gray!20}\textbf{0.7323} \\
     & CLIPIQA+$\uparrow$ & 0.6222 & 0.6360 & 0.7171 & 0.7241 & 0.5094 & 0.7136 & \underline{0.7292} & 0.7111 & 0.5888 & \cellcolor{gray!20}\textbf{0.7388} \\
     & NIMA$\uparrow$ & 5.1866 & 5.2607 & 5.4154 & \underline{5.4648} & 5.0419 & 5.3530 & 5.3920 & \textbf{5.5045} & 5.1763 & \cellcolor{gray!20}5.3212 \\
     & NIQE$\downarrow$ & 4.1796 & 4.3515 & 3.9996 & \textbf{3.7803} & 4.8432 & \underline{3.9268} & 3.9991 & 3.9936 & 5.3042 & \cellcolor{gray!20}4.0156 \\
     & LIQE$\uparrow$ & 3.4836 & 3.4319 & \underline{4.1806} & 4.0541 & 2.6938 & 3.9967 & 4.1690 & 4.0778 & 3.2175 & \cellcolor{gray!20}\textbf{4.3648} \\
     & MUSIQ$\uparrow$ & 62.9605 & 63.3346 & 70.2502 & 68.7120 & 57.4580 & 69.5654 & \underline{70.8834} & 68.9061 & 61.3634 & \cellcolor{gray!20}\textbf{71.3676} \\
     & MANIQA$\uparrow$ & 0.5553 & 0.5749 & 0.6360 & 0.6385 & 0.4925 & 0.6143 & 0.6418 & \underline{0.6481} & 0.5343 & \cellcolor{gray!20}\textbf{0.6545} \\
    \bottomrule
  \end{tabular}
  }
\end{table*}

\subsubsection{Quantitative Comparisons}
\begin{figure*}[t!]
    \centering
    \includegraphics[width=0.96\linewidth]{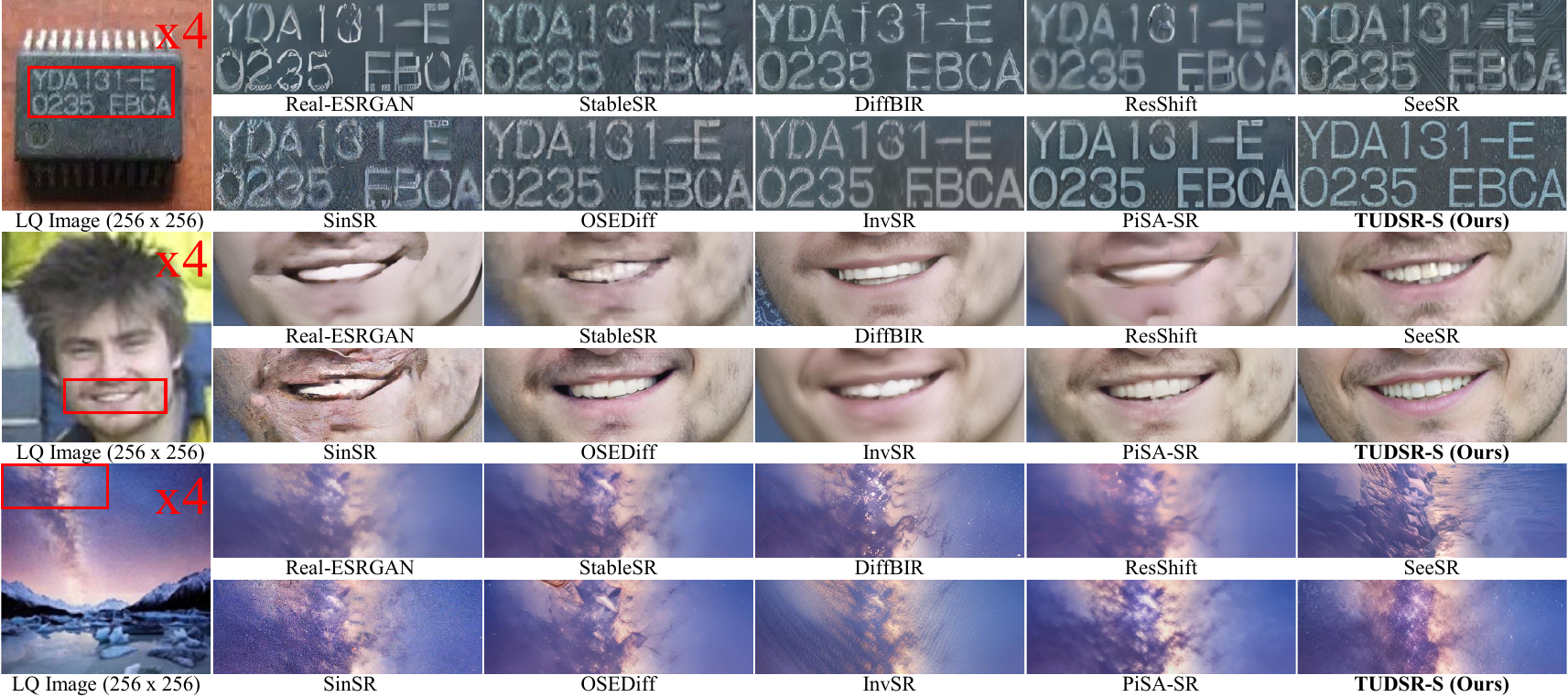}
    \caption{Qualitative comparisons ($\times4$ \ie $256^2\rightarrow{1024^2}$) with state-of-the-art multi-step and one-step models. Please \textbf{zoom in}.}
    \vspace{-2mm}
    \label{fig:cmp4}
\end{figure*}

\begin{figure*}[t!]
    \centering
    \includegraphics[width=0.96\linewidth]{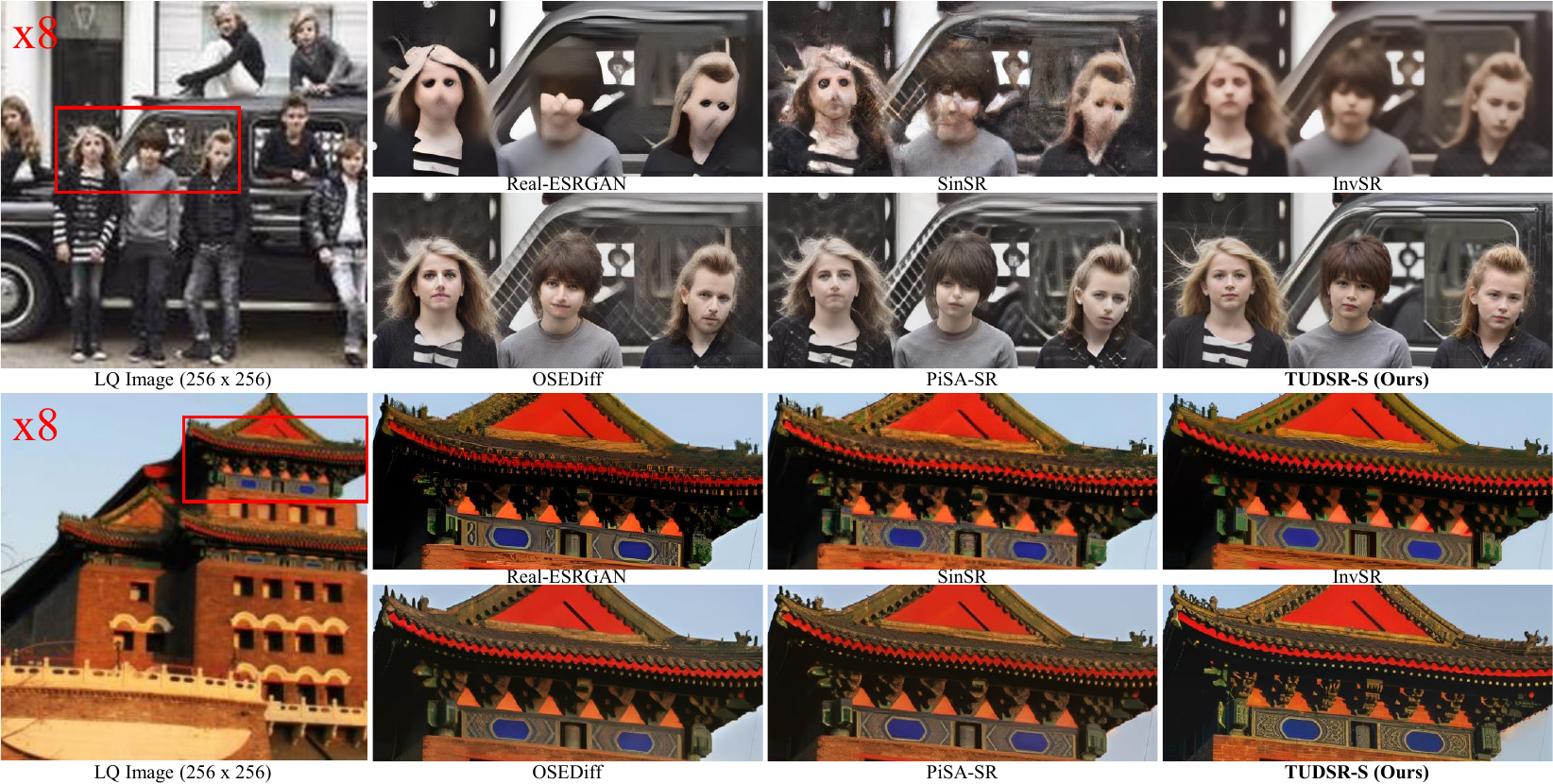}
    \caption{Qualitative comparisons ($\times8$ \ie $256^2\rightarrow{2048^2}$) with state-of-the-art one-step models. Please \textbf{zoom in}.}
    \vspace{-2mm}
    \label{fig:cmp8}
\end{figure*}
\Cref{tab:cmp4} presents the comprehensive quantitative comparisons ($\times 4$) on four test datasets. Our proposed TUDSR-S demonstrates state-of-the-art performance across multiple real-world SR benchmarks, which achieves overwhelming results on key perceptual quality metrics, including CLIPIQA, CLIPIQA+, LIQE, MUSIQ, and MANIQA. Such metrics consistently reflect the quality of images generated by TUDSR-S from various perspectives.

\Cref{tab:cmp8} also presents the quantitative comparisons ($\times 8$) to demonstrate the quality of our generation on high-resolution (\ie $1024^2$ and $2048^2$) images. 
The one-step models exhibit a substantial decline across most evaluation metrics on the $\times 8$ SR tests conducted on the RealSR, DRealSR, and RealLQ250 datasets, compared to \cref{tab:cmp4}. This indicates that the $\times 8$ upsampling task exceeds the capacity of these one-step approaches. In contrast, TUDSR-S achieves comprehensive performance across most metrics across the three datasets. The results in both \cref{tab:cmp4} and \cref{tab:cmp8} demonstrate that our method not only delivers strong performance in the conventional $\times 4$ setting but also excels in the more challenging $\times 8$ SR scenario, showing the effectiveness of our twice upsampling–diffusion strategy.

\begin{table}
    \caption{Quantitative comparisons ($\times8$) with state-of-the-art one-step models on four real-world benchmark datasets. Note that TUDSR-S denotes \texttt{M4N2} twice upsampling-diffusion in the table.}
    \vspace{-3mm}
    \label{tab:cmp8}
    \resizebox{\columnwidth}{!}{
    \centering
    \begin{tabular}{c|c|ccccccc}
    \toprule
    Datasets & Methods & C-IQA$\uparrow$ & C-IQA+$\uparrow$ & NIMA$\uparrow$ & NIQE$\downarrow$ & LIQE$\uparrow$ & MUSIQ$\uparrow$ & MANIQA$\uparrow$
    \\
    \midrule
    \multirow{6}{*}{RealSR}
    & RealESRGAN & 0.5061 & 0.5833 & 4.9421 & 5.3701 & 3.2344 & 57.6718 & 0.4858 \\
    & OSEDiff & \underline{0.6976} & \underline{0.6673} & \textbf{5.0711} & \underline{5.6951} & \underline{3.6347} & \textbf{67.5970} & 0.5678 \\
    & PiSA-SR & 0.6642 & 0.6562 & 4.8645 & 5.0937 & 3.2765 & 66.0150 & 0.5510 \\
    & InvSR & 0.6581 & 0.6420 & \underline{4.9595} & \textbf{4.3930} & 3.0830 & 64.3007 & \underline{0.5711} \\
    & SinSR & 0.6175 & 0.5151 & 4.6538 & 7.2585 & 2.4228 & 53.4213 & 0.4605 \\
    \rowcolor{gray!20}
    & TUDSR-S & \textbf{0.6920} & \textbf{0.6883} & 4.8305 & 4.6839 & \textbf{3.6547} & \underline{67.2225} & \textbf{0.6126} \\
    \midrule
    \multirow{6}{*}{DrealSR}
    & RealESRGAN & 0.4642 & 0.5073 & 4.5156 & 6.8732 & 2.4309 & 47.5704 & 0.4226 \\
    & OSEDiff & \underline{0.6774} & \underline{0.6265} & \underline{4.6819} & \underline{6.1817} & \underline{2.8764} & \underline{58.5390} & \underline{0.5176} \\
    & PiSA-SR & 0.6572 & 0.6084 & 4.5354 & 6.2137 & 2.7529 & 56.7244 & 0.4960 \\
    & InvSR & 0.5621 & 0.5640 & 4.6442 & \textbf{5.2770} & 2.6230 & 54.4117 & 0.5028 \\
    & SinSR & 0.6118 & 0.5051 & 4.4498 & 7.8526 & 2.3197 & 48.7678 & 0.4236 \\
    \rowcolor{gray!20}
    & TUDSR-S & \textbf{0.7186} & \textbf{0.6680} & \textbf{4.7217} & 5.5417 & \textbf{3.6124} & \textbf{63.9383} & \textbf{0.5758} \\
    \midrule
    \multirow{6}{*}{RealLQ250}
    & RealESRGAN & 0.5370 & 0.5629 & \underline{4.9598} & \underline{4.6685} & 2.7754 & 48.7098 & 0.4549 \\
    & OSEDiff & \underline{0.6793} & \underline{0.6288} & 4.7853 & 4.6349 & 2.7375 & \underline{54.8024} & \underline{0.5171} \\
    & PiSA-SR & 0.6355 & 0.6110 & 4.7429 & 4.7085 & 2.4839 & 49.8787 & 0.4801 \\
    & InvSR & 0.5233 & 0.5057 & 4.8383 & 5.5685 & 2.2872 & 40.2874 & 0.4358 \\
    & SinSR & 0.6222 & 0.4930 & 4.7463 & 5.7477 & 1.9833 & 40.1009 & 0.4362 \\
    \rowcolor{gray!20}
    & TUDSR-S & \textbf{0.7222} & \textbf{0.6904} & \textbf{5.3103} & \textbf{3.9251} & \textbf{3.6112} & \textbf{63.2129} & \textbf{0.5847} \\
    \bottomrule
    \end{tabular}
    }
\end{table}

\subsubsection{Qualitative Comparisons}
We present qualitative comparisons for $\times4$ super-resolution in \cref{fig:cmp4}. In the first case (top row), our TUDSR-S produces the best quality for characters and numbers, followed by OSEDiff, InvSR, and PiSA-SR, while the other models perform relatively poorly. In the second case, our method achieves the highest quality in generating teeth and beard details, with the beard appearance being particularly realistic. OSEDiff, PiSA-SR, and SeeSR show lower realism and clarity than our model, while the remaining models exhibit significant quality degradation (especially SinSR and ResShift). In the third case, TUDSR-S generates the most realistic starry sky, complete with fine star details. DiffBIR also produces reasonable results, though inferior to ours, and other models perform noticeably worse.

We further provide qualitative comparisons for $\times8$ SR in \cref{fig:cmp8}. These examples demonstrate the strong performance of our TUDSR-S in high-resolution generation. In the first case (top row), our approach produces a facial image closest to the real person, with rich details and a naturally rendered bokeh background of the vehicle. Real-ESRGAN, SinSR, and InvSR yield the poorest results with missing details, while OSEDiff and PiSA-SR generate faces with inconsistent identity. The second case highlights TUDSR-S’s superiority in fine-detail generation, such as roof eaves, where our model surpasses all others. Competing models produce blurry, oversmoothed results in these regions.

These comparisons fully demonstrate that TUDSR-S achieves strong performance on both standard $\times4$ and challenging $\times8$ SR tasks, validating the effectiveness of our twice upsampling-diffusion strategy for generating high-fidelity details. We consider that the two diffusion processes can fully utilize the powerful prior of SD.

\subsubsection{Inference Time Comparisons}
\begin{table}[t]
    \label{tab:inference_time}
    \centering
    \caption{
    Inference time (seconds per image). All inference times are tested on an H800 GPU.
    }
    \vspace{-3mm}
    \resizebox{\columnwidth}{!}{
    \begin{tabular}{lccccc}
        \toprule
        SR task & SinSR & OSEDiff & InvSR & PiSA-SR & TUDSR-S \\
        \midrule
        $\times4$ ($128^2\rightarrow512^2$) & \underline{0.0728} & 0.0892& 0.0838 & 0.0789 & \textbf{0.0596}\\
        \midrule
        $\times8$ ($128^2\rightarrow1024^2$) & 0.5674 & 0.6532& \textbf{0.3693} & 0.5601 & \underline{0.4201}\\
        \midrule
        $\times4$ ($256^2\rightarrow1024^2$) & 0.5889 & 0.6776 & \textbf{0.3723} & 0.5776 & \underline{0.4296}\\
        \midrule
        $\times8$ ($256^2\rightarrow2048^2$) & 3.2941 & 3.9241 & \textbf{1.7844} & 2.8337 & \underline{2.2848}  \\
        \bottomrule
    \end{tabular}
    }
    \label{tab:tb2}
\end{table}
We also provide the inference time in~\cref{tab:inference_time}. On the low-resolution task ($128^2\rightarrow512^2$), TUDSR-S exhibits the fastest inference speed ($0.0596$ s), significantly outperforming other methods.
On high-resolution tasks (\eg, $256^2\rightarrow2048^2$), InvSR achieves the best performance ($1.7844$ s) and demonstrates a significant advantage over other methods.
Overall, TUDSR-S and InvSR alternately lead in tasks with varying resolutions, while OSEDiff consistently takes the longest time across all tested scenarios.

\subsection{Ablation Study}
Since our TUDSR-S is a twice upsampling-diffusion model, it involves two-stage upsampling during inference. Thus, we conduct ablation studies on TUDSR-S, primarily focusing on $\times4$ and $\times8$ SR tasks, as presented in ~\cref{tab:up4,tab:up8}. Note that in the tables, \texttt{M4}/\texttt{N4} denotes $M=4$/$N=4$ using only the stage-1/stage-2 LoRA SR model. Similarly, \texttt{M8}/\texttt{N8} denotes $M=4$/$N=4$. \texttt{M2N2} indicates $M=2$ using the first LoRA SR model, and then $N=2$ using the second LoRA SR model. Likewise, \texttt{M4N2} follows suit, and so on. Please refer to~\cref{fig:arch}(b) for a better understanding.

\subsubsection{Ablation Study on TUDSR-S ($\times4$)}
In the $\times4$ SR task, the resolutions obtained after $\times4$ upsampling RealSR, DRealSR, and RealLQ250 are $512^2$, $512^2$, and $1024^2$, respectively. RealLR200 has non-fixed resolutions, and after $\times4$ upsampling, the resulting resolution ranges from $512$ to $1728$. As shown in \cref{tab:up4}, \texttt{M2N2} achieves the best overall performance across all metrics and datasets. Meanwhile, \texttt{M4}, which follows the conventional $\times4$ SR approach, performs second best overall performance. In contrast, \texttt{N4} yields extremely poor results. The experiments demonstrate that our twice upsampling-diffusion method, which decomposes $\times4$ SR into two $\times2$ upsampling stages, also achieves outstanding performance.

\subsubsection{Ablation Study on TUDSR-S ($\times8$)}
In the $\times8$ SR task, the resolutions obtained after $\times8$ upsampling RealSR, DRealSR, and RealLQ250 are $1024^2$, $1024^2$, and $2048^2$, respectively. In \cref{tab:up8}, \texttt{M4N2} significantly outperforms \texttt{M8} and \texttt{N8} across all metrics on RealLQ250. On RealSR and DRealSR, \texttt{M4N2} also achieves the best overall performance. Similarly, \texttt{N8} produces extremely poor results. This validates the effectiveness of our twice upsampling-diffusion approach, demonstrating that decomposing $\times8$ SR into two stages ($\times4$ and $\times2$) is an effective strategy for achieving higher SR.
\begin{figure}[t]
    \centering
    \includegraphics[width=1\linewidth]{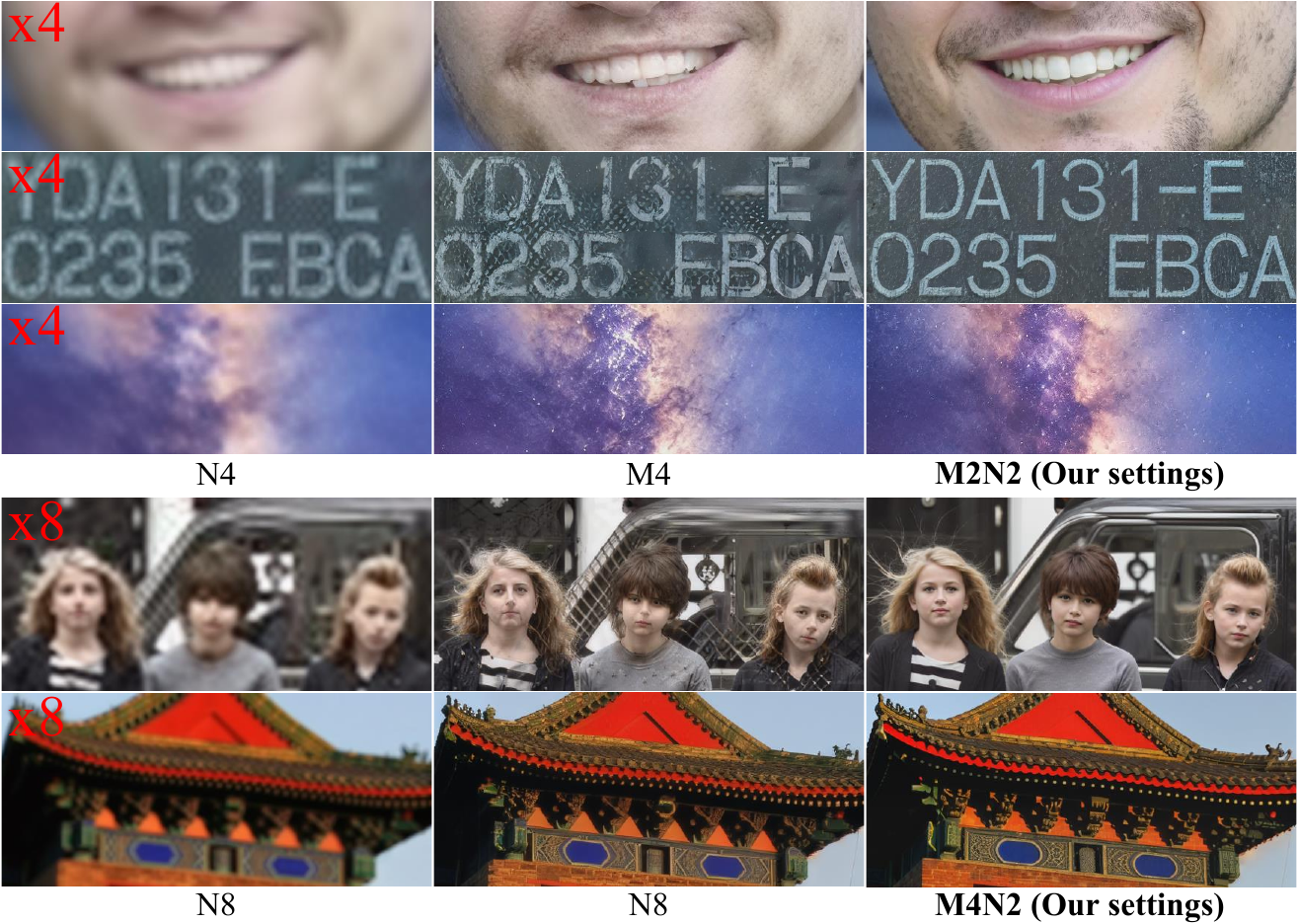}
    \vspace{-2mm}
    \caption{Visualization of twice upsampling-diffusion ($\times4$/$\times8$ \ie $256^2\rightarrow{1024^2}$/$256^2\rightarrow{2048^2}$). Please \textbf{zoom in}.}
    \label{fig:ablation}
\end{figure}
\subsubsection{Visualization on TUDSR-S ($\times4$ and $\times8$)}

\Cref{fig:ablation} provides a visualization of the aforementioned ablation study. In both $\times4$ and $\times8$ SR tasks, \texttt{M2N2} and \texttt{M4N2} demonstrate outstanding performance in terms of detail reproduction. In comparison, \texttt{M4} and \texttt{M8} exhibit relatively fewer details, while \texttt{N4} and \texttt{N8} fail to generate satisfactory results. The quality of the generated images further demonstrates the effectiveness of our method.

\begin{table}[t]
\caption{Ablation study on twice upsampling-diffusion (total $\times 4$).}
\vspace{-3mm}
\label{tab:up4}
\resizebox{\columnwidth}{!}{
\centering
\begin{tabular}{c|c|ccccccc}
\toprule
    Datasets & Type & C-IQA$\uparrow$ & C-IQA+$\uparrow$ & NIMA$\uparrow$ & NIQE$\downarrow$ & LIQE$\uparrow$ & MUSIQ$\uparrow$ & MANIQA$\uparrow$ \\
    \midrule
    \multirow{3}{*}{RealSR}
    & \texttt{M4} & \underline{0.6657} & \underline{0.6952} & \underline{4.8549} & \underline{5.0086} & \underline{4.1895} & \underline{69.2568} & \textbf{0.6818} \\
    & \texttt{N4} & 0.3056 & 0.4397 & 3.8149 & 8.2005 & 1.3971 & 28.5303 & 0.3472 \\
    \rowcolor{gray!20}
    & \texttt{M2N2} & \textbf{0.6846} & \textbf{0.7135} & \textbf{4.8676} & \textbf{4.7149} & \textbf{4.3738} & \textbf{70.2406} & \underline{0.6786} \\
    \midrule
    \multirow{3}{*}{DrealSR}
    & \texttt{M4} & \textbf{0.6984} & \underline{0.6905} & \underline{4.7232} & \underline{5.6939} & \underline{4.1050} & \underline{65.7759} & \textbf{0.6387} \\
    & \texttt{N4} & 0.2947 & 0.3790 & 3.5315 & 9.8559 & 1.2533 & 23.9991 & 0.3269 \\
    \rowcolor{gray!20}
    & \texttt{M2N2} & \underline{0.6944} & \textbf{0.6921} & \textbf{4.7272} & \textbf{5.5427} & \textbf{4.1769} & \textbf{66.3650} & \underline{0.6312} \\
    \midrule
    \multirow{3}{*}{RealLQ250}
    & \texttt{M4} & \underline{0.7124} & \underline{0.7177} & \underline{5.2862} & \textbf{3.5207} & \underline{4.1257} & \underline{70.4876} & \textbf{0.6351} \\
    & \texttt{N4} & 0.2953 & 0.3939 & 4.1043 & 7.4385 & 1.3163 & 31.3223 & 0.3341 \\
    \rowcolor{gray!20}
    & \texttt{M2N2} & \textbf{0.7315} & \textbf{0.7297} & \textbf{5.3268} & \underline{3.9440} & \textbf{4.3017} & \textbf{71.9250} & \underline{0.6329} \\
    \midrule
    \multirow{3}{*}{RealLR200}
    & \texttt{M4} & \underline{0.7109} & \underline{0.7290} & \textbf{5.3926} & \textbf{3.6845} & \underline{4.2609} & \underline{70.3921} & \textbf{0.6611} \\
    & \texttt{N4} & 0.3677 & 0.4853 & \underline{4.4882} & \underline{5.4621} & 1.7350 & 42.7009 & 0.4200 \\
    \rowcolor{gray!20}
    & \texttt{M2N2} & \textbf{0.7323} & \textbf{0.7388} & \underline{5.3212} & 4.0156 & \textbf{4.3648} & \textbf{71.3676} & \underline{0.6545} \\
    \bottomrule
    \end{tabular}
    }
\end{table}

\begin{table}[t]
\caption{Ablation study on twice upsampling-diffusion (total $\times 8$).}
\vspace{-3mm}
\label{tab:up8}
\resizebox{\columnwidth}{!}{
\centering
\begin{tabular}{c|c|ccccccc}
\toprule
    Datasets & Type & C-IQA$\uparrow$ & C-IQA+$\uparrow$ & NIMA$\uparrow$ & NIQE$\downarrow$ & LIQE$\uparrow$ & MUSIQ$\uparrow$ & MANIQA$\uparrow$ \\
    \midrule
    \multirow{3}{*}{RealSR}
    & \texttt{M8} & \textbf{0.6928} & \underline{0.6833} & \textbf{4.8379} & \textbf{4.4895} & \underline{3.3738} & \underline{65.7316} & \underline{0.5970} \\
    & \texttt{N8} & 0.2672 & 0.2933 & 3.8038 & 11.0800 & 1.1892 & 19.6518 & 0.3013 \\
    \rowcolor{gray!20}
    & \texttt{M4N2} & \underline{0.6920} & \textbf{0.6883} & \underline{4.8305} & \underline{4.6839} & \textbf{3.6547} & \textbf{67.2225} & \textbf{0.6126} \\
    \midrule
    \multirow{3}{*}{DrealSR}
    & \texttt{M8} & \underline{0.7056} & \underline{0.6554} & \underline{4.6973} & \textbf{5.3613} & \underline{3.1561} & \underline{60.1592} & \underline{0.5490} \\
    & \texttt{N8} & 0.2987 & 0.2738 & 3.5905 & 12.9729 & 1.3113 & 19.5022 & 0.3086 \\
    \rowcolor{gray!20}
    & \texttt{M4N2} & \textbf{0.7186} & \textbf{0.6680} & \textbf{4.7217} & \underline{5.5417} & \textbf{3.6124} & \textbf{63.9383} & \textbf{0.5758} \\
    \midrule
    \multirow{3}{*}{RealLQ250}
    & \texttt{M8} & \underline{0.6723} & \underline{0.6347} & \underline{4.5271} & \underline{4.1017} & \underline{2.7134} & \underline{51.5389} & \underline{0.5309} \\
    & \texttt{N8} & 0.2635 & 0.2712 & 3.9308 & 11.1885 & 1.1709 & 19.6954 & 0.2946 \\
    \rowcolor{gray!20} 
    & \texttt{M4N2} & \textbf{0.7222} & \textbf{0.6904} & \textbf{5.3103} & \textbf{3.9251} & \textbf{3.6112} & \textbf{63.2129} & \textbf{0.5847} \\
    \bottomrule
    \end{tabular}
    }
\end{table}

\section{Conclusion}
In this work, we present TUDSR, a novel Twice Upsampling–Diffusion SR framework that effectively addresses the challenge of high-resolution image generation from a low-resolution generative model. By decomposing the demanding upsampling process into two manageable stages, our approach successfully circumvents the resolution limitations inherent in native SD models. Based on SD2.1-base, we instantiate TUDSR-S, which achieves excellent performance in $\times4$ and $\times8$ SR tasks, especially in addressing the shortcomings of existing SR models in the high-resolution SR tasks (\eg $2048^2$).
Furthermore, the TUDSR framework establishes a generalizable strategy, paving the way for its application to larger generative models (\eg FLUX.1-dev) and future advancements towards $4096^2$ resolution. 

\section{Acknowledgments}
This research is supported by the General Program of Shanghai Natural Science Foundation (No.24ZR1419800, No.23ZR1419300), the National Natural Science Foundation of China (No.42130112), the Ministry of Industry and Information Technology of China, Science and Technology Commission of Shanghai Municipality (No.22DZ2229004), and the Shanghai Frontiers Science Center of Molecule Intelligent Syntheses.
{
    \small
    \bibliographystyle{ieeenat_fullname}
    \bibliography{main}
}

\clearpage
\setcounter{page}{1}
\maketitlesupplementary
\section{More Visualizations on TUDSR-S ($\times8$)}
\begin{figure*}
    \centering
    \includegraphics[width=0.90\linewidth]{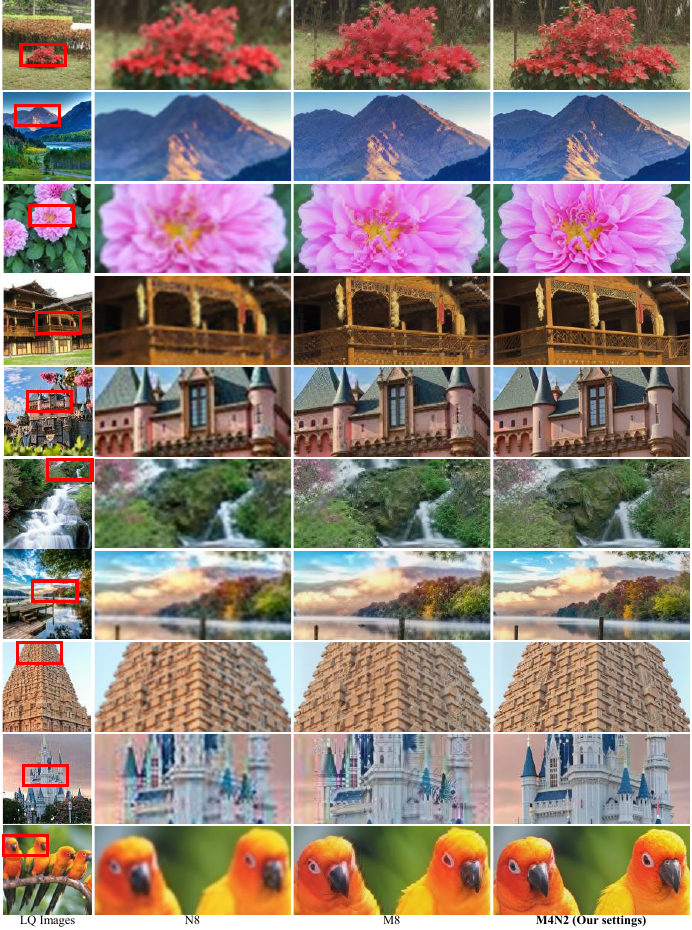}
    \caption{Visualization of twice upsampling-diffusion ($\times8$ \ie $256^2\rightarrow2048^2$). The LQ images (from top to bottom) are from RealLQ250 (014, 080, 100, 104, 111, 154, 166, 212, 228, 230). Please \textbf{zoom in}.}
    \label{fig:more_ablation}
\end{figure*}

\Cref{fig:more_ablation} shows more visualizations of twice upsampling-diffusion ($\times8$) on TUDSR-S. \texttt{M4N2} achieves the best clarity and detail across all $10$ cases from RealLQ250. The quality of the image generated by \texttt{M8} is also significantly lower than that of \texttt{M4N2}, while \texttt{N8} achieves the worst results. This result indicates that decomposing $\times8$ into $\times4$ and $\times2$ has excellent performance in high-resolution generation.

\section{More Qualitative Comparisons ($\times8$)}
\begin{figure*}
    \centering
    \includegraphics[width=0.97\linewidth]{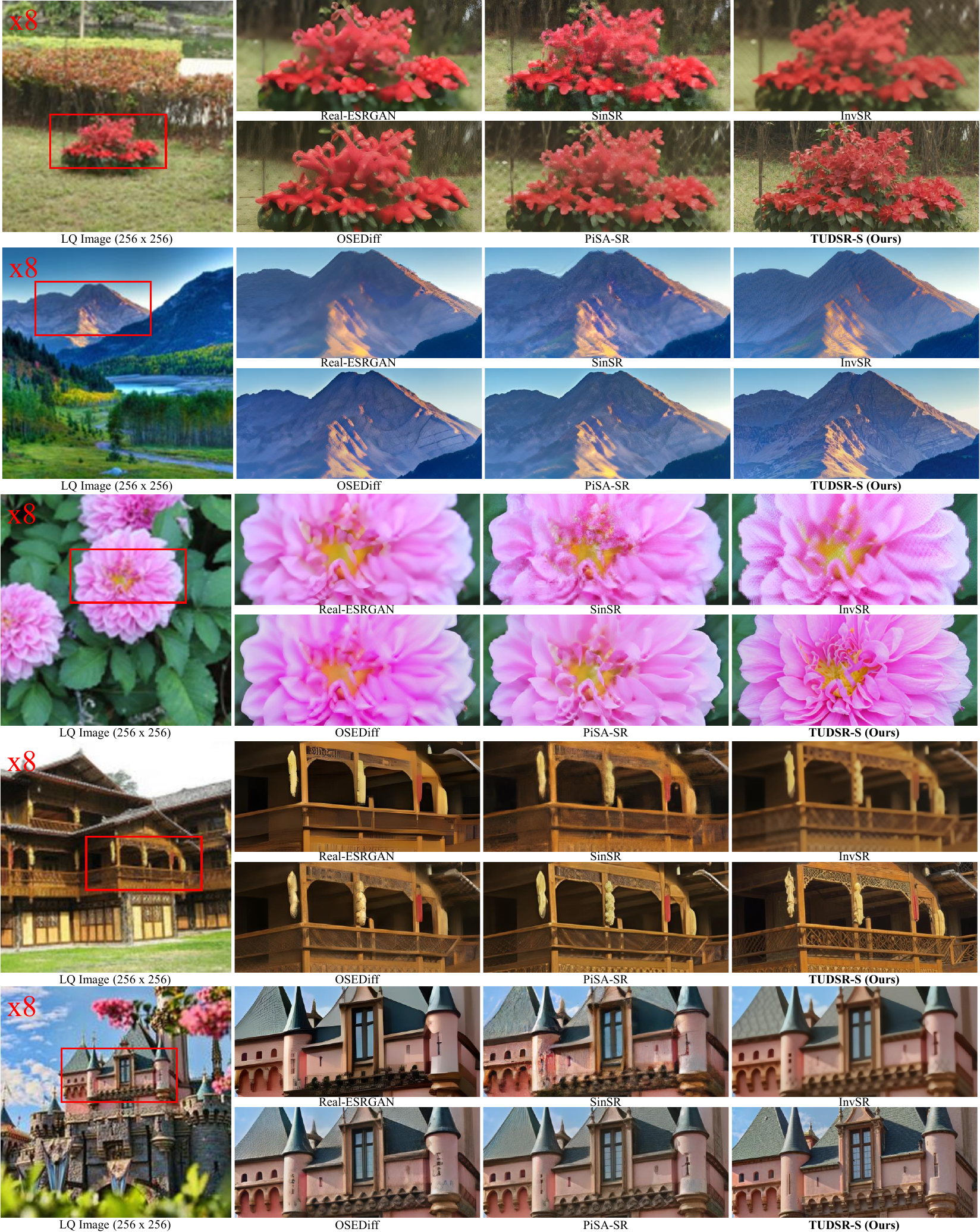}
    \caption{Qualitative comparisons ($\times8$ \ie $256^2\rightarrow 2048^2$) with state-of-the-art one-step models. The LQ images (from top to bottom) are from RealLQ250 (014, 080, 100, 104, 111). Please \textbf{zoom in}.}
    \label{fig:more_cmp8}
\end{figure*}

\begin{figure*}
    \centering
    \includegraphics[width=0.97\linewidth]{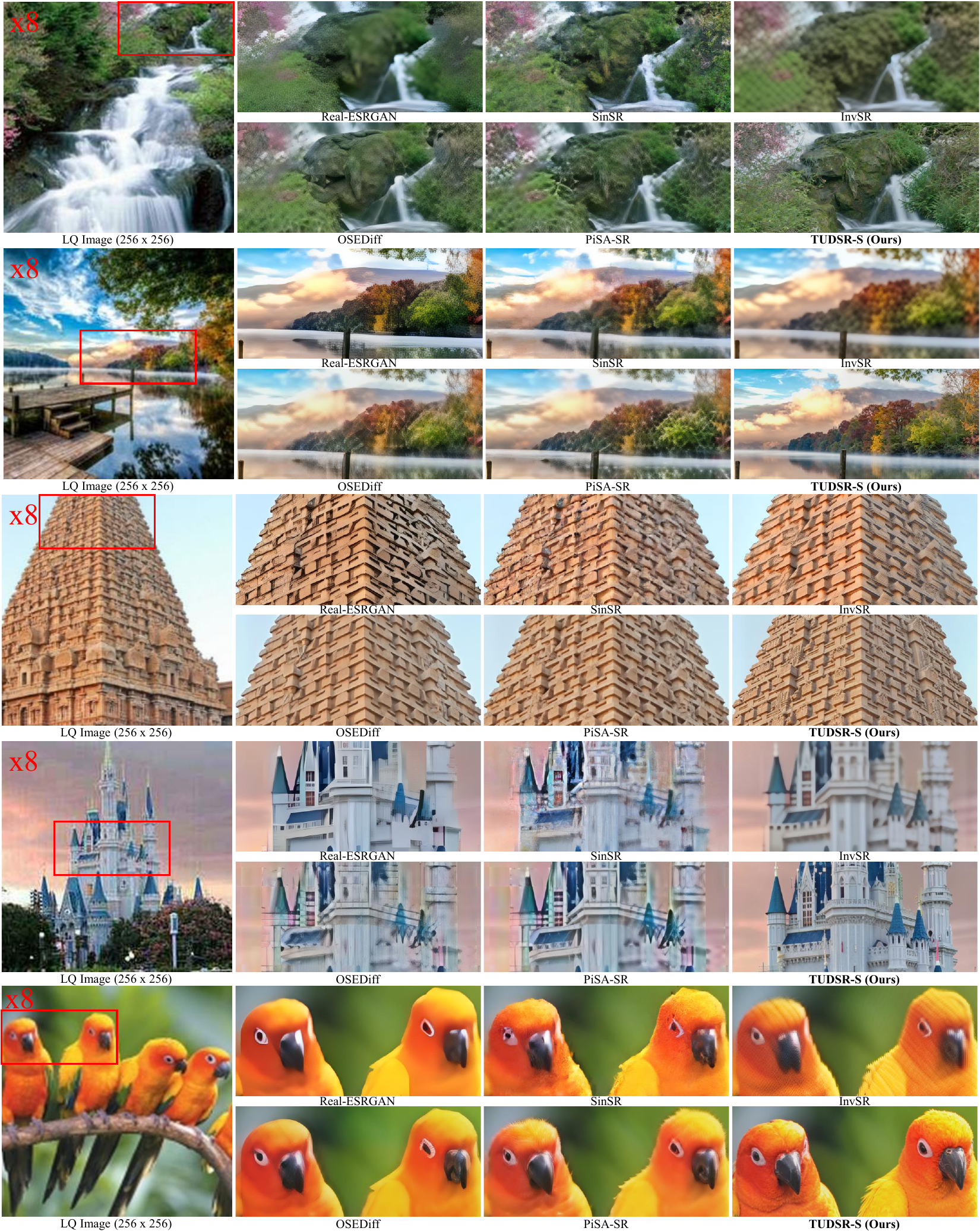}
    \caption{Qualitative comparisons ($\times8$ \ie $256^2\rightarrow 2048^2$) with state-of-the-art one-step models. The LQ images (from top to bottom) are from RealLQ250 (154, 166, 212, 228, 230). Please \textbf{zoom in}.}
    \label{fig:more_cmp8-2}
\end{figure*}

\Cref{fig:more_cmp8,fig:more_cmp8-2} show more visual comparisons of $\times 8$ SR ($256^2\rightarrow2048^2$). TUDSR-S exhibits overwhelming performance across these one-step models, highlighting the effectiveness of our twice upsampling-diffusion method.

\end{document}